\documentclass{article}

\usepackage{graphicx}

\usepackage[final]{neurips_2025_ml4ps}

\usepackage[utf8]{inputenc} 
\usepackage[T1]{fontenc}    
\usepackage{hyperref}       
\usepackage{url}            
\usepackage{booktabs}       
\usepackage{amsfonts}       
\usepackage{nicefrac}       
\usepackage{microtype}      
\usepackage{xcolor}         
\usepackage{amsmath}
\usepackage{booktabs}
\usepackage{siunitx}
\usepackage{tabularx}

\title{Active Learning for Machine Learning Driven Molecular Dynamics}

\author{%
  Kevin Bachelor\\
  University of California, Santa Cruz\\
  \texttt{kwbachel@ucsc.edu} \\
  \And
  Sanya Murdeshwar \\
  University of California, Santa Cruz\\
  \texttt{smurdesh@ucsc.edu} \\
  \AND
  Daniel Sabo \\
  University of California, Santa Cruz\\
  \texttt{dsabo@ucsc.edu} \\
  \And
  Ashwin Lokapally \\
  GiwoTech Inc. \\
  \texttt{ashwin@giwotech.com} \\
  \And
  Razvan Marinescu \\
  University of California, Santa Cruz\\
  \texttt{ramarine@ucsc.edu} \\
}

\begin{document}

\maketitle

\begin{abstract}
    Machine-learned coarse-grained (CG) potentials are fast, but degrade over time when simulations reach under-sampled bio-molecular conformations, and generating widespread all-atom (AA) data to combat this is computationally infeasible. We propose a novel active learning (AL) framework for CG neural network potentials in molecular dynamics (MD). Building on the CGSchNet model~\cite{Schnet}, our method employs root mean squared deviation (RMSD)-based frame selection from MD simulations in order to generate data on-the-fly by querying an oracle during the training of a neural network potential. This framework preserves CG-level efficiency while correcting the model at precise, RMSD-identified coverage gaps. By training CGSchNet—a coarse-grained neural network potential—we empirically show that our framework  explores previously unseen configurations and trains the model on unexplored regions of conformational space. Our active learning framework enables a CGSchNet model trained on the Chignolin protein to achieve a 33.05\% improvement in the Wasserstein-1 (W1) metric in Time-lagged Independent Component Analysis (TICA)~\cite{Noe2015} space on an in-house benchmark suite.

\end{abstract}

\section{Introduction}
Advancements in the realm of MD have allowed for the effective exploration of complex conformational landscapes of biomolecular structures such as proteins, but such methods remain limited by the immense computational overhead associated with system size and the granularity of the atomic representation~\cite{hollingsworth2018molecular}. Multiple generations of AA neural network potentials have progressed high-resolution MD by taking atomic coordinates, ${\mathbf{r} = (\mathbf{r}_1, ..., \mathbf{r}_N)} \in \mathbb{R}^{3N}$, and predicting the energy $E = U_{\theta}(\mathbf{r})$ with forces $\mathbf{F}(\mathbf{r}) = -\nabla_{\mathbf{r}} U_{\theta}(\mathbf{r})$, surpassing classical energy functionals in terms of accuracy~\cite{Behler2021, Duignan2024, Sumpter_Noid_1992, Unke_Meuwly_2019, Bartók_Payne_Kondor_Csányi_2010, Behler_2011, Behler_Parrinello_2007}. However, such high-dimensional models remain slow at AA resolution and are often more computationally expensive than computing classical force-field terms.~\cite{hollingsworth2018molecular} Consequently, simulating large structures takes a prohibitive amount of time to reach critical and relevant regions, thereby making speedup and efficiency key goals in MD development.

A renowned approach for achieving longer timescales with neural network potentials in MD is to reduce the size of the conformational space through coarse-graining, in which groups of atoms are represented as single beads~\cite{Kmiecik_Gront_Kolinski_Wieteska_Dawid_Kolinski_2016}. Generally, such models have been trained via force matching, which fits the CG potential to match reference AA forces on mapped configurations to minimize the difference between CG forces and mapped AA forces~\cite{Husic2020}. However, because most training data samples single metastable states in proteins, they struggle to generalize to unseen states and fully explore the conformational space, particularly during transitions between metastable states, making simulations prone to conformational "explosion" or "implosion" anomalies. These divergences result from the network generating physically inconsistent forces upon encountering configurations significantly different from those in the initial training data. This issue is especially prevalent for large proteins with complex folding structures. 

To address this, we propose an active learning framework that uses targeted, on-the-fly data acquisition to obtain data beyond the regions normally sampled by MD by selecting specific out-of-distribution frames during simulation and querying an AA ground-truth oracle, thereby patching coverage gaps at minimal cost. For data acquisition, we identify frames most different from the training set quantified by distance-based proxies such as RMSD, enabling data generation in the least-explored regions of the protein conformational space. We evaluate our work using  a recently-published benchmark for AI-driven dynamics \cite{Aghili2025}, which uses a free-energy matching loss in TICA space. Our active learning framework achieves an improvement of 33.05\% in TICA space compared to a standard training setup without AL. In addition, we find the AL framework results in significant qualitative exploration of the conformational space as visualized in the TICA plots. 

\subsection{Related Works}

Previous approaches to improve data efficiency in MD simulations have involved both active learning strategies as well as CG neural network potentials, but these avenues have largely been explored separately. On the active learning front, the Deep Potential Generator (DP-GEN) serves as an early active learning framework for AA neural network potentials for materials simulations~\cite{PhysRevMaterials.3.023804}. More recently, a framework proposed by Jung et al. introduces another active learning strategy for AA neural network potentials, demonstrating the benefits of uncertainty-guided sampling in minimizing redundant data generation during molecular simulations~\cite{D3DD00216K}. These frameworks, however, were designed for AA systems and do not address the CG perspective, which inherently imposes additional limitations. More recently, Duschatko et al.~\cite{Duschatko_Vandermause_Molinari_Kozinsky_2024} propose a Bayesian uncertainty-aware active learning framework that does address CG free energy models, in which CG configurations with high uncertainty trigger active learning, though this workflow does not clearly refer to a full all-atom oracle for every sampled configuration. 

In parallel, coarse-graining has long been used to increase simulation timescales by reducing model resolution~\cite{Husic2020, Kmiecik_Gront_Kolinski_Wieteska_Dawid_Kolinski_2016, Marrink_Risselada_Yefimov_Tieleman_deVries_2007, Wang_Olsson_Wehmeyer_Pérez_Charron_deFabritiis_Noé_Clementi_2019}. Classical top-down CG methods later emerged such as the Martini force field, reproducing experimental observations through parametric model tuning~\cite{Marrink_Risselada_Yefimov_Tieleman_deVries_2007}. In contrast, bottom-up CG approaches derive CG potentials from AA simulations to retain physical accuracy, usually by employing force matching~\cite{Jin_Pak_Durumeric_Loose_Voth_2022}. In recent years, machine learning has been applied to bottom-up CG approaches with notable success. For instance, Wang et al.~\cite{Wang_Olsson_Wehmeyer_Pérez_Charron_deFabritiis_Noé_Clementi_2019} demonstrates that a neural network (CGNet) is able to learn a CG free energy surface via force matching, though their framework requires manual feature selection for molecules. With this foundation, developments such as CGSchNet~\cite{Husic2020}, building upon the SchNet model~\cite{Schnet}, have significantly advanced coarse-graining via a graph neural network architecture, achieving strong thermodynamical accuracy. However, such models are trained on fixed datasets and therefore lack a mechanism for adaptive data acquisition, leading to problematic behavior when simulations enter under-sampled regions of conformational space.

\section{Methodology}

\begin{figure}
    \centering
    \includegraphics[width=0.7\linewidth]{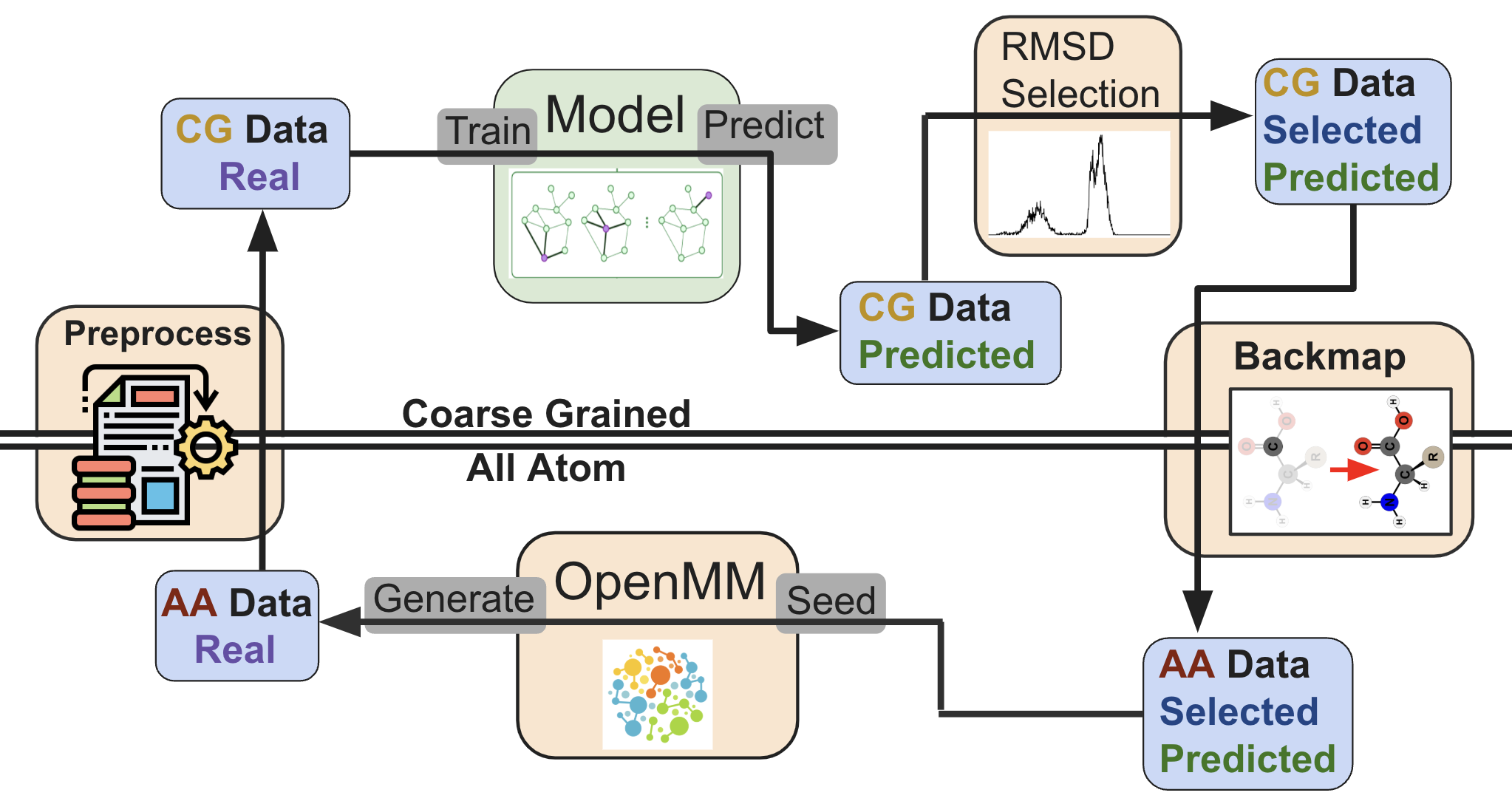}
    \caption{Our active learning pipeline showing the model-training $\Leftrightarrow$ data-generation loop. Each round trains a coarse-grained (CG) model, runs a CG simulation, selects least-covered frames via RMSD distance proxy, backmaps them to all-atom (AA) space, runs short OpenMM simulations, and projects AA$\to$CG to augment the dataset and retrain. By querying the oracle only where coverage is poor, the loop increases conformational coverage at minimal AA simulation cost.}
    \label{fig:pipeline}
\end{figure}

In Fig.~\ref{fig:pipeline} we show an overview of our active learning methodology. At each active learning iteration, the model is trained on MD trajectory data and then used to simulate the CG protein system. We then select the frames with the largest RMSD discrepancies, indicating the greatest difference between the configuration and the training dataset, ultimately targeting the least-covered configurations. These frames are then backmapped into AA space, used to query the oracle by seeding OpenMM~\cite{eastman2023openmm} all-atom simulations, projected back into CG space, and appended to the dataset. The model is retrained, and the loop repeats, querying the oracle only where coverage is low to minimize usage of AA simulations.

\subsection{Neural network potential (CGSchNet)}

Our neural network leverages a graph neural network (GNN) architecture based on CGSchNet~\cite{Schnet}, using continuous filter convolutions to build features based on inter-bead distances. Let us define the CG bead coordinates as $\mathbf{R} = (\mathbf{R}_1, ..., \mathbf{R}_M)\in \mathbb{R}^{3M}$, and the inter-bead distance as $r_{ij} = ||\mathbf{R}_i - \mathbf{R}_j||$. Based on these distances, the network outputs a scalar energy potential $\mathbf{U_{\theta}(\mathbf{R}})$ which accounts for the system's geometry rather than just the absolute position or orientation. This architecture thereby ensures that $\mathbf{U_{\theta}}$ remains invariant to global translations and rotations, and the resultant forces $\mathbf{F_{\theta}(\mathbf{R}) = -\nabla_{\mathbf{R}} U_{\theta}(\mathbf{R}) }$ are translation-invariant and rotation-equivariant as well. The objective of force matching is done by minimizing the mean-squared error between the predicted forces and the target CG forces projected from AA space. The force matching loss function $\mathcal{L_{FM}}(\theta)$ is given as: 

\[
\mathcal{L_{FM}}(\theta)
= \frac{1}{T}\sum_{t=1}^{T}
\frac{1}{3M^{(t)}}\left\|
\mathbf F_\theta\!\left(\mathbf R^{(t)}\right) - \mathbf F_{\mathrm{CG}}^{(t)}
\right\|_F^{2}
\]

where $\mathbf R^{(t)} \in \mathbb{R}^{3M^{(t)}}$ represents the CG bead coordinates for $M^{(t)}$ beads in frame $t$, $F_{\mathrm{CG}}^{(t)}$ represents the AA$\rightarrow$CG projected CG forces and $T$ tells us the number of training frames.

\subsection{AA $\rightarrow$ CG and CG $\rightarrow$ AA projection}
Our neural network potential is compatible with CG data, while the oracle simulator works only with AA data. Thus, we require a bidirectional bridge between the spaces. We convert AA to CG by mapping atomic coordinates to carbon-alpha ($C\alpha$) beads via a pre-processing script using a linear operator
\[
\Xi:\,\mathbb{R}^{3N}\to\mathbb{R}^{3M}, \qquad \mathbf{R} = \Xi\,\mathbf{r}.
\]
We map the forces by projecting AA forces onto CG degrees of freedom via force projection operators~\cite{Schnet, Husic2020}:
\begin{gather*}
\mathbf{F}_{\mathrm{CG}} = \Xi_F\,\mathbf{f}_{\mathrm{AA}} \\
\Xi_F = (\Xi \Xi^\top)^{-1}\,\Xi
\end{gather*}
To convert from CG to AA, we utilize the PULCHRA~\cite{rotkiewicz2008} backmapper to reconstruct the missing, non-$C\alpha$ atoms into statistically likely positions.

\subsection{Frame selection}
For frame selection, we use the RMSD distance proxy for concrete visualization of coverage gaps. To select frames, during each active learning iteration we compute histograms of the RMSD values between simulated frames and the training set.  We then select the frames with the largest simulation-minus-training RMSD discrepancies as candidates, signifying abnormal gaps in coverage. We then filter further by removing any frames with RMSD value outside of a cutoff to avoid getting frames that are part of implosions or explosions. These frames are then backmapped to AA space in order to seed the OpenMM~\cite{eastman2023openmm} oracle which generates robust data that is then projected back into CG space to begin retraining.

For readers unfamiliar with high-dimensional stochastic data, it may seem that a distance-based acquisition function would bias selection toward later parts of the trajectory, since the simulation begins in-distribution and would generally drift outwards. In practice, this does not occur. Trajectories typically spend long intervals fluctuating around local minima before occasionally jumping to new basins, where they again remain for many steps. A newly-reached RMSD region may be previously explored or be entirely new. As a result, variance in the projected metric does not grow linearly with time, and RMSD-based acquisition does not inherently favor later frames.
\section{Results}
To test the efficacy of our active learning framework, we trained a base CGSchnet model on a small subset of data from the Chignolin protein, and used an in-house benchmark suite~\cite{Aghili2025} to measure the (i) conformation exploration along slow modes of motion via TICA projections~\cite{Noe2015}, (ii) reaction coordinates representing the fraction of time certain atoms are within close proximity, and the distributions of (iii) bond lengths, (iv) bond angles, and (v) dihedral angles. Model-generated distributions were compared against extensive Chignolin ground-truth references, depicted in Figure~\ref{fig:pre_compare} for the base model, and in Figure~\ref{fig:post_compare} for the model after applying the active learning loop. We summarize the outcomes quantitatively with the W1 distance metric between distributions, representing the minimal "work" required to transform the model distribution into the reference, where lower values are favorable. In Table  ~\ref{tab:al-metrics} we detail the results of the CGSchnet\cite{Husic2020} model before applying the active learning, the same model after our active learning have been applied, as well as another recent model that utilizes a metalearning framework for a Chignolin trained CGSchnet model as reference\cite{aghili2025ticabasedfreeenergymatching}. The table details the W1 distance between the ground truth, and each model's simulations, for the distributions over the 5 measurements we took for Chignolin.

\begin{table}[th]
\centering
\caption{Wasserstein-1 (W1) before and after active learning (lower is better).}
\label{tab:al-metrics}
\begin{tabular}{lccl}
\toprule
 & \multicolumn{3}{c}{W1$\downarrow$}\\
\cmidrule(lr){2-4}
Plot& Before AL & \textbf{After AL}&Match Energy\cite{aghili2025ticabasedfreeenergymatching}\\
\midrule
TICA KDE            & 1.15023 & \textbf{0.77003}  &1.17342\\
Reaction coordinate & \textbf{0.15141} & 0.38302  &0.65711\\
Bond length         & 0.00043 & \textbf{0.00022}  &0.00057\\
Bond angle          & 0.11036 & \textbf{0.10148}  &0.24846\\
Dihedral            & 0.25472& 0.36378  &\textbf{0.21482}\\
\bottomrule
\end{tabular}
\end{table}

\begin{figure}
    \centering
    \includegraphics[width=0.9\linewidth]{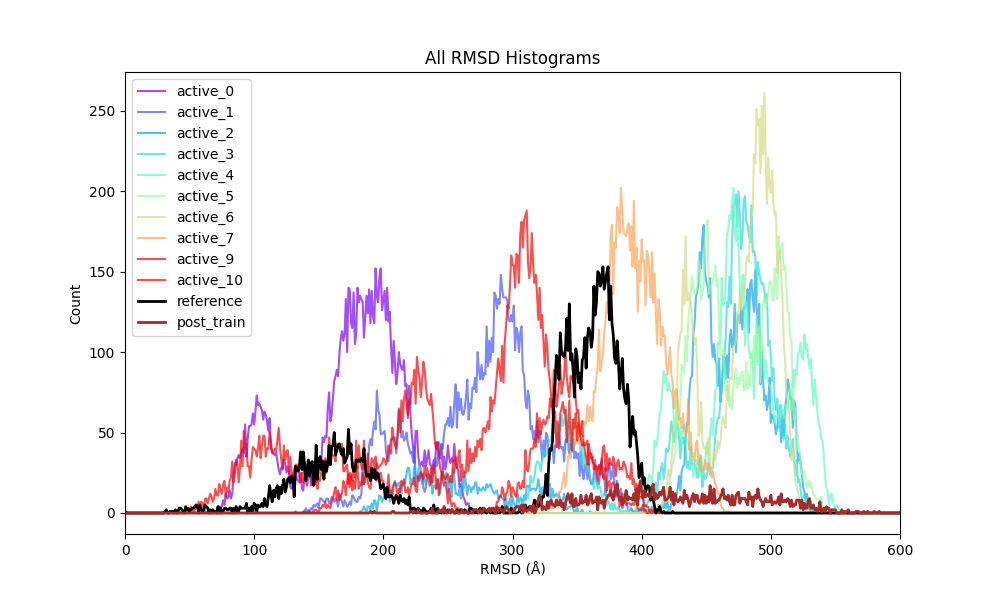}
    \caption{Base model output (black), model output after active learning (brown), and training data per active learning iteration (purple $\rightarrow$ red), plotted as a histogram of RMSD values from the reference frame.}
    \label{fig:histograms}
\end{figure}

These benchmarks show significant improvements in exploration capability and local accuracy measures. In TICA space, the W1 distance to the reference dropped by 33.05\%. Additionally, it produced more stable trajectories, with bond-length and bond-angle distributions aligning more closely with the ground truth with a 48.84\% and 8.05\% decrease in W1 distance, respectively.

The dihedral and RC metrics did not improve, which is expected: the AL dihedral benchmark graph -- shown in \ref{fig:post_compare} -- visually matches well, given that dihedral angles are inherently noisy. Furthermore, the RC metric captures the PDF of a single atom-pair distance, representing a highly localized, magnified view of the configuration space, and therefore is extremely sensitive to global changes. These localized deviations do not contradict the broader trend, as the other metrics that track global conformational structure all show strong improvement.

Improvements in exploration are also evident in the RMSD histograms before and after active learning (Figure~\ref{fig:histograms}). The base model (black) shows a bimodal RMSD density distribution, indicating sampling concentrated in two conformational states. The additional histograms represent the active learning iterations (purple $\rightarrow$ red), reflecting the data that was used to train the model at each iteration. The substantial variance across iterations indicates that each round, the model targets different under-sampled regions, with the active learning loop supplying training data to support it across diverse conformational states. The post-active learning model (brown) shows a much broader RMSD distribution, further illustrating improvements in phase-space coverage aligning with the TICA W1 reductions reported above.

\section{Discussion}
Since we query OpenMM only when the RMSD histograms indicate poorly covered regions, we retain the speed and efficiency of CG simulations while correcting the model in those under-sampled states on-the-fly, restoring accuracy and thoroughness in coverage. Evaluated on Chignolin's complex structure, pure CG simulation without using our active learning framework often resulted in explosions or implosions due to limited coverage and indicative of poor generalization over the conformational space of the protein. Empirically, we found that our active learning method explored more of the conformational space and improved accuracy across the bond, bond-angle, dihedral, and TICA space metrics in the standardized benchmark. This indicates that $(i)$ long CG trajectories are stabilized and explosion and implosion anomalies are avoided, and $(ii)$ more of the conformation space is explored, which is crucial for the ultimate goal of generalization. 

These results demonstrate that targeted active learning in MD stabilizes CG simulations and expands conformational coverage within reasonable compute, while also providing paths for future improvements upon mechanisms of our framework that impose limitations on its efficacy. Stronger backmapping methodologies would allow for increased accuracy in AA reconstructions while reducing relaxation cost. Furthermore, more rigorous distance proxies or measures may further optimize how frames are prioritized and, in turn, reduce excessive oracle usage. Ultimately, our active learning framework's targeted, precise, oracle-backed patching of phase-space gaps can significantly expand the training data for ML models in molecular dynamics at a practical cost, and provides clear paths for improvements in selection, backmapping, and representation bridging.

\section*{Impact statement}
Accurate yet affordable exploration of protein conformational spaces is a critical bottleneck in ML-driven drug discovery, where rare conformational states and transitions govern crucial protein mechanisms such as binding ability. Our active learning framework unifies high-speed CG neural network potentials with precise all-atom oracle queries to explore conformational spaces more rigorously while keeping the computational cost of labeling low. Our result enables increased capability in viewing protein behavior, which may play a role in revealing unique binding opportunities or promising compounds far sooner, minimizing the time spent in extensive trial-and-error in drug discovery laboratories. Because our framework is model-agnostic and bridges CG$\leftrightarrow$AA reliably, it can be used to augment current or future state-of-the-art ML force fields easily rather than replacing them altogether, resulting in confident observations and shorter paths from trial-and-error to effective treatments.

\bibliographystyle{plain}
\bibliography{references}

@article{Husic2020,
  title = {Coarse graining molecular dynamics with graph neural networks},
  volume = {153},
  ISSN = {1089-7690},
  url = {http://dx.doi.org/10.1063/5.0026133},
  DOI = {10.1063/5.0026133},
  number = {19},
  journal = {The Journal of Chemical Physics},
  publisher = {AIP Publishing},
  author = {Husic,  Brooke E. and Charron,  Nicholas E. and Lemm,  Dominik and Wang,  Jiang and Pérez,  Adrià and Majewski,  Maciej and Kr\"{a}mer,  Andreas and Chen,  Yaoyi and Olsson,  Simon and de Fabritiis,  Gianni and Noé,  Frank and Clementi,  Cecilia},
  year = {2020},
  month = nov 
}

@misc{aghili2025ticabasedfreeenergymatching,
      title={TICA-Based Free Energy Matching for Machine-Learned Molecular Dynamics}, 
      author={Alexander Aghili and Andy Bruce and Daniel Sabo and Razvan Marinescu},
      year={2025},
      eprint={2509.14600},
      archivePrefix={arXiv},
      primaryClass={cs.LG},
      url={https://arxiv.org/abs/2509.14600}, 
}

@article{Aghili2025,
    title = {A Standardized Benchmark for Machine-Learned Molecular Dynamics Using Weighted Ensemble Sampling},
    author = {Aghili, Alexander and Bruce, Andy and Sabo, Daniel and Murdeshwar, Sanya and Bachelor, Kevin and Mistreanu, Ionut and Lokapally, Ashwin and Marinescu, Razvan},
    volume = {129},
    ISSN = {1520-5207},
    url = {http://dx.doi.org/10.1021/acs.jpcb.5c05365},
    DOI = {10.1021/acs.jpcb.5c05365},
    number = {50},
    journal = {The Journal of Physical Chemistry B},
    publisher = {American Chemical Society (ACS)},
    year = {2025},
    month = Dec,
    pages = {12828–12840}
}

@article{Behler2021,
  title = {Four Generations of High-Dimensional Neural Network Potentials},
  volume = {121},
  PMID = {33779150},
  url = {https://doi.org/10.1021/acs.chemrev.0c00868},
  DOI = {10.1021/acs.chemrev.0c00868},
  number = {16},
  journal = {Chemical Reviews},
  publisher = {American Chemical Society},
  author = {Behler, J{\"o}rg},
  year = {2021},
  month = march 
}

@inproceedings{Schnet,
author = {Sch\"{u}tt, K. T. and Kindermans, P.-J. and Sauceda, H. E. and Chmiela, S. and Tkatchenko, A. and M\"{u}ller, K.-R.},
title = {SchNet: a continuous-filter convolutional neural network for modeling quantum interactions},
year = {2017},
isbn = {9781510860964},
publisher = {Curran Associates Inc.},
address = {Red Hook, NY, USA},
booktitle = {Proceedings of the 31st International Conference on Neural Information Processing Systems},
pages = {992–1002},
numpages = {11},
location = {Long Beach, California, USA},
series = {NIPS'17}
}

@article{Noe2015,
    author = {No{\'e}, Frank and Clementi, Cecilia},
    title = {Kinetic Distance and Kinetic Maps from Molecular Dynamics Simulation},
    journal = {Journal of Chemical Theory and Computation},
    volume = {11},
    number = {10},
    pages = {5002-5011},
    year = {2015},
    doi = {10.1021/acs.jctc.5b00553},
        note ={PMID: 26574285},
    URL = { 
            https://doi.org/10.1021/acs.jctc.5b00553
    },
    eprint = {     
            https://doi.org/10.1021/acs.jctc.5b00553
    }
}

@article{rotkiewicz2008,
    author = {Rotkiewicz, Piotr and Skolnick, Jeffrey},
    title = {Fast procedure for reconstruction of full-atom protein models from reduced representations},
    journal = {Journal of Computational Chemistry},
    year = {2008},
    volume = {29},
    number = {9},
    pages = {1460--1465},
    month = {July},
    doi = {10.1002/jcc.20906},
    pmid = {18196502},
    pmcid = {PMC2692024}
}

@article{eastman2023openmm,
    title={OpenMM 8: Molecular Dynamics Simulation with Machine Learning Potentials},
    author={Eastman, Peter and Galvelis, Raimondas and Peláez, Raúl P. and Abreu, Charlles R. A. and Farr, Sarah E. and Gallicchio, Emilio and Gorenko, Anton and Henry, Matthew M. and Hu, Feng and Huang, Jing and Krämer, Andreas and Michel, Julien and Mitchell, Joshua A. and Pande, Vijay S. and Rodrigues, João P. G. L. M. and Rodriguez-Guerra, Jaime and Simmonett, Andrew C. and Singh, Sukrit and Swails, Jason and Turner, Peter and Wang, Yuanqing and Zhang, Ivy and Chodera, John D. and De Fabritiis, Gianni and Markland, Thomas E.},
    journal={J. Phys. Chem. B},
    volume={128},
    number={1},
    pages={109--116},
    year={2023},
    publisher={American Chemical Society}
}

@article{hollingsworth2018molecular,
    title={Molecular Dynamics Simulation for All},
    author={Hollingsworth, Scott A. and Dror, Ron O.},
    journal={Neuron},
    volume={99},
    number={6},
    pages={1129--1143},
    year={2018},
    publisher={Elsevier},
    doi={10.1016/j.neuron.2018.08.011},
    url={https://doi.org/10.1016/j.neuron.2018.08.011},
    issn={0896-6273}
}

@article{Duignan2024,
author = {Duignan, Timothy T.},
title = {The Potential of Neural Network Potentials},
journal = {ACS Physical Chemistry Au},
volume = {4},
number = {3},
pages = {232-241},
year = {2024},
doi = {10.1021/acsphyschemau.4c00004},

URL = { 
    
        https://doi.org/10.1021/acsphyschemau.4c00004
    
    

},
eprint = { 
    
        https://doi.org/10.1021/acsphyschemau.4c00004
    
    

}

}

@article{D3DD00216K,
author ="Jung, Gang Seob and Choi, Jong Youl and Lee, Sangkeun Matthew",
title  ="Active learning of neural network potentials for rare events",
journal  ="Digital Discovery",
year  ="2024",
volume  ="3",
issue  ="3",
pages  ="514-527",
publisher  ="RSC",
doi  ="10.1039/D3DD00216K",
url  ="http://dx.doi.org/10.1039/D3DD00216K"}

@article{Duschatko_Vandermause_Molinari_Kozinsky_2024, title={Uncertainty driven active learning of coarse grained free energy models}, volume={10}, rights={2024 The Author(s)}, ISSN={2057-3960}, DOI={10.1038/s41524-023-01183-5}, abstractNote={Coarse graining techniques play an essential role in accelerating molecular simulations of systems with large length and time scales. Theoretically grounded bottom-up models are appealing due to their thermodynamic consistency with the underlying all-atom models. In this direction, machine learning approaches hold great promise to fitting complex many-body data. However, training models may require collection of large amounts of expensive data. Moreover, quantifying trained model accuracy is challenging, especially in cases of non-trivial free energy configurations, where training data may be sparse. We demonstrate a path towards uncertainty-aware models of coarse grained free energy surfaces. Specifically, we show that principled Bayesian model uncertainty allows for efficient data collection through an on-the-fly active learning framework and opens the possibility of adaptive transfer of models across different chemical systems. Uncertainties also characterize models’ accuracy of free energy predictions, even when training is performed only on forces. This work helps pave the way towards efficient autonomous training of reliable and uncertainty aware many-body machine learned coarse grain models.}, number={1}, journal={npj Computational Materials}, publisher={Nature Publishing Group}, author={Duschatko, Blake R. and Vandermause, Jonathan and Molinari, Nicola and Kozinsky, Boris}, year={2024}, month=jan, pages={9}, language={en} }

@article{PhysRevMaterials.3.023804,
  title = {Active learning of uniformly accurate interatomic potentials for materials simulation},
  author = {Zhang, Linfeng and Lin, De-Ye and Wang, Han and Car, Roberto and E, Weinan},
  journal = {Phys. Rev. Mater.},
  volume = {3},
  issue = {2},
  pages = {023804},
  numpages = {9},
  year = {2019},
  month = {Feb},
  publisher = {American Physical Society},
  doi = {10.1103/PhysRevMaterials.3.023804},
  url = {https://link.aps.org/doi/10.1103/PhysRevMaterials.3.023804}
}

@article{Marrink_Risselada_Yefimov_Tieleman_deVries_2007, title={The MARTINI Force Field: Coarse Grained Model for Biomolecular Simulations}, volume={111}, ISSN={1520-6106}, DOI={10.1021/jp071097f}, abstractNote={We present an improved and extended version of our coarse grained lipid model. The new version, coined the MARTINI force field, is parametrized in a systematic way, based on the reproduction of partitioning free energies between polar and apolar phases of a large number of chemical compounds. To reproduce the free energies of these chemical building blocks, the number of possible interaction levels of the coarse-grained sites has increased compared to those of the previous model. Application of the new model to lipid bilayers shows an improved behavior in terms of the stress profile across the bilayer and the tendency to form pores. An extension of the force field now also allows the simulation of planar (ring) compounds, including sterols. Application to a bilayer/cholesterol system at various concentrations shows the typical cholesterol condensation effect similar to that observed in all atom representations.}, number={27}, journal={The Journal of Physical Chemistry B}, publisher={American Chemical Society}, author={Marrink, Siewert J. and Risselada, H. Jelger and Yefimov, Serge and Tieleman, D. Peter and de Vries, Alex H.}, year={2007}, month=july, pages={7812–7824} }

@article{Kmiecik_Gront_Kolinski_Wieteska_Dawid_Kolinski_2016, title={Coarse-Grained Protein Models and Their Applications}, volume={116}, ISSN={0009-2665}, DOI={10.1021/acs.chemrev.6b00163}, abstractNote={The traditional computational modeling of protein structure, dynamics, and interactions remains difficult for many protein systems. It is mostly due to the size of protein conformational spaces and required simulation time scales that are still too large to be studied in atomistic detail. Lowering the level of protein representation from all-atom to coarse-grained opens up new possibilities for studying protein systems. In this review we provide an overview of coarse-grained models focusing on their design, including choices of representation, models of energy functions, sampling of conformational space, and applications in the modeling of protein structure, dynamics, and interactions. A more detailed description is given for applications of coarse-grained models suitable for efficient combinations with all-atom simulations in multiscale modeling strategies.}, number={14}, journal={Chemical Reviews}, publisher={American Chemical Society}, author={Kmiecik, Sebastian and Gront, Dominik and Kolinski, Michal and Wieteska, Lukasz and Dawid, Aleksandra Elzbieta and Kolinski, Andrzej}, year={2016}, month=july, pages={7898–7936} }

@article{Wang_Olsson_Wehmeyer_Pérez_Charron_deFabritiis_Noé_Clementi_2019, title={Machine Learning of Coarse-Grained Molecular Dynamics Force Fields}, volume={5}, ISSN={2374-7943}, DOI={10.1021/acscentsci.8b00913}, abstractNote={Atomistic or ab initio molecular dynamics simulations are widely used to predict thermodynamics and kinetics and relate them to molecular structure. A common approach to go beyond the time- and length-scales accessible with such computationally expensive simulations is the definition of coarse-grained molecular models. Existing coarse-graining approaches define an effective interaction potential to match defined properties of high-resolution models or experimental data. In this paper, we reformulate coarse-graining as a supervised machine learning problem. We use statistical learning theory to decompose the coarse-graining error and cross-validation to select and compare the performance of different models. We introduce CGnets, a deep learning approach, that learns coarse-grained free energy functions and can be trained by a force-matching scheme. CGnets maintain all physically relevant invariances and allow one to incorporate prior physics knowledge to avoid sampling of unphysical structures. We show that CGnets can capture all-atom explicit-solvent free energy surfaces with models using only a few coarse-grained beads and no solvent, while classical coarse-graining methods fail to capture crucial features of the free energy surface. Thus, CGnets are able to capture multibody terms that emerge from the dimensionality reduction.}, number={5}, journal={ACS Central Science}, publisher={American Chemical Society}, author={Wang, Jiang and Olsson, Simon and Wehmeyer, Christoph and Pérez, Adrià and Charron, Nicholas E. and de Fabritiis, Gianni and Noé, Frank and Clementi, Cecilia}, year={2019}, month=may, pages={755–767} }

@article{Jin_Pak_Durumeric_Loose_Voth_2022, title={Bottom-up Coarse-Graining: Principles and Perspectives}, volume={18}, ISSN={1549-9618}, DOI={10.1021/acs.jctc.2c00643}, abstractNote={Large-scale computational molecular models provide scientists a means to investigate the effect of microscopic details on emergent mesoscopic behavior. Elucidating the relationship between variations on the molecular scale and macroscopic observable properties facilitates an understanding of the molecular interactions driving the properties of real world materials and complex systems (e.g., those found in biology, chemistry, and materials science). As a result, discovering an explicit, systematic connection between microscopic nature and emergent mesoscopic behavior is a fundamental goal for this type of investigation. The molecular forces critical to driving the behavior of complex heterogeneous systems are often unclear. More problematically, simulations of representative model systems are often prohibitively expensive from both spatial and temporal perspectives, impeding straightforward investigations over possible hypotheses characterizing molecular behavior. While the reduction in resolution of a study, such as moving from an atomistic simulation to that of the resolution of large coarse-grained (CG) groups of atoms, can partially ameliorate the cost of individual simulations, the relationship between the proposed microscopic details and this intermediate resolution is nontrivial and presents new obstacles to study. Small portions of these complex systems can be realistically simulated. Alone, these smaller simulations likely do not provide insight into collectively emergent behavior. However, by proposing that the driving forces in both smaller and larger systems (containing many related copies of the smaller system) have an explicit connection, systematic bottom-up CG techniques can be used to transfer CG hypotheses discovered using a smaller scale system to a larger system of primary interest. The proposed connection between different CG systems is prescribed by (i) the CG representation (mapping) and (ii) the functional form and parameters used to represent the CG energetics, which approximate potentials of mean force (PMFs). As a result, the design of CG methods that facilitate a variety of physically relevant representations, approximations, and force fields is critical to moving the frontier of systematic CG forward. Crucially, the proposed connection between the system used for parametrization and the system of interest is orthogonal to the optimization used to approximate the potential of mean force present in all systematic CG methods. The empirical efficacy of machine learning techniques on a variety of tasks provides strong motivation to consider these approaches for approximating the PMF and analyzing these approximations.}, number={10}, journal={Journal of Chemical Theory and Computation}, publisher={American Chemical Society}, author={Jin, Jaehyeok and Pak, Alexander J. and Durumeric, Aleksander E. P. and Loose, Timothy D. and Voth, Gregory A.}, year={2022}, month=oct, pages={5759–5791} }

@article{Sumpter_Noid_1992, title={Potential energy surfaces for macromolecules. A neural network technique}, volume={192}, ISSN={0009-2614}, DOI={10.1016/0009-2614(92)85498-Y}, abstractNote={A method for obtaining potential energy surfaces for macromolecules is described. The basis of the method is the use of a neural network to learn the relationship between vibrational spectra and a multidimensional potential energy surface (PES). The results demonstrate that the neural network is capable of mapping the vibrational motion determined from spectra onto a fully coupled PES with relatively high levels of accuracy.}, number={5}, journal={Chemical Physics Letters}, author={Sumpter, Bobby G. and Noid, Donald W.}, year={1992}, month=may, pages={455–462} }

@article{Bartók_Payne_Kondor_Csányi_2010, title={Gaussian Approximation Potentials: The Accuracy of Quantum Mechanics, without the Electrons}, volume={104}, DOI={10.1103/PhysRevLett.104.136403}, abstractNote={We introduce a class of interatomic potential models that can be automatically generated from data consisting of the energies and forces experienced by atoms, as derived from quantum mechanical calculations. The models do not have a fixed functional form and hence are capable of modeling complex potential energy landscapes. They are systematically improvable with more data. We apply the method to bulk crystals, and test it by calculating properties at high temperatures. Using the interatomic potential to generate the long molecular dynamics trajectories required for such calculations saves orders of magnitude in computational cost.}, number={13}, journal={Physical Review Letters}, publisher={American Physical Society}, author={Bartók, Albert P. and Payne, Mike C. and Kondor, Risi and Csányi, Gábor}, year={2010}, month=apr, pages={136403} }

@article{Behler_2011, title={Neural network potential-energy surfaces in chemistry: a tool for large-scale simulations}, volume={13}, ISSN={1463-9084}, DOI={10.1039/C1CP21668F}, abstractNote={The accuracy of the results obtained in molecular dynamics or Monte Carlo simulations crucially depends on a reliable description of the atomic interactions. A large variety of efficient potentials has been proposed in the literature, but often the optimum functional form is difficult to find and strongly depends on the particular system. In recent years, artificial neural networks (NN) have become a promising new method to construct potentials for a wide range of systems. They offer a number of advantages: they are very general and applicable to systems as different as small molecules, semiconductors and metals; they are numerically very accurate and fast to evaluate; and they can be constructed using any electronic structure method. Significant progress has been made in recent years and a number of successful applications demonstrate the capabilities of neural network potentials. In this Perspective, the current status of NN potentials is reviewed, and their advantages and limitations are discussed.}, number={40}, journal={Physical Chemistry Chemical Physics}, publisher={The Royal Society of Chemistry}, author={Behler, Jörg}, year={2011}, month=oct, pages={17930–17955}, language={en} }

@article{Behler_Parrinello_2007, title={Generalized Neural-Network Representation of High-Dimensional Potential-Energy Surfaces}, volume={98}, DOI={10.1103/PhysRevLett.98.146401}, abstractNote={The accurate description of chemical processes often requires the use of computationally demanding methods like density-functional theory (DFT), making long simulations of large systems unfeasible. In this Letter we introduce a new kind of neural-network representation of DFT potential-energy surfaces, which provides the energy and forces as a function of all atomic positions in systems of arbitrary size and is several orders of magnitude faster than DFT. The high accuracy of the method is demonstrated for bulk silicon and compared with empirical potentials and DFT. The method is general and can be applied to all types of periodic and nonperiodic systems.}, number={14}, journal={Physical Review Letters}, publisher={American Physical Society}, author={Behler, Jörg and Parrinello, Michele}, year={2007}, month=apr, pages={146401} }





\newpage
\appendix
\renewcommand{\thefigure}{A.\arabic{figure}}
\setcounter{figure}{0}

\section{Technical Appendices and Supplementary Material}

\begin{figure}[h]
    \centering
    \includegraphics[width=1.0\textwidth]{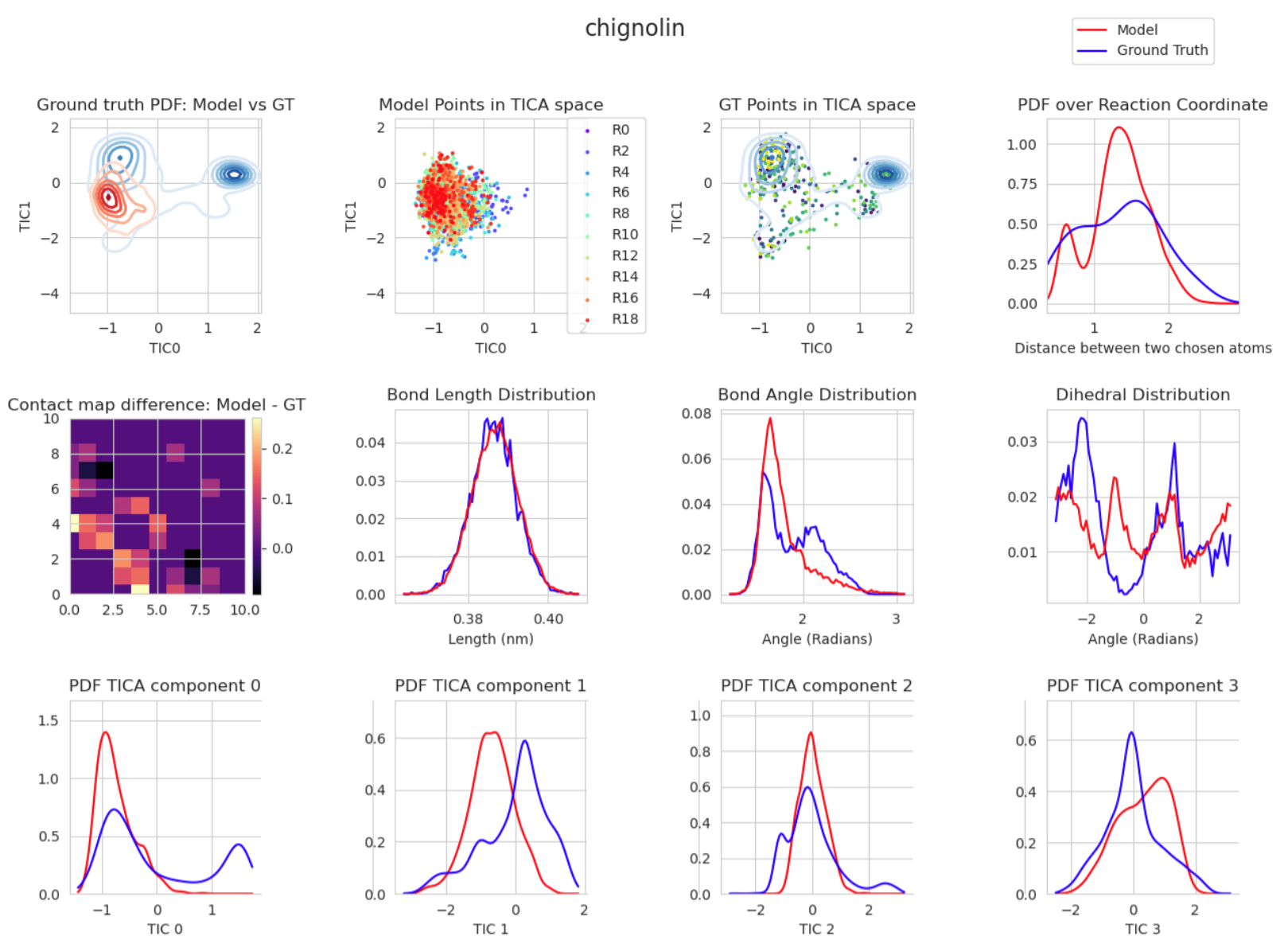}    
    \caption{Benchmark suite evaluating the base model's performance before applying the active learning loop.}
    \label{fig:pre_compare}
\end{figure}
\begin{figure}[h]
    \centering
    \includegraphics[width=1.0\textwidth]{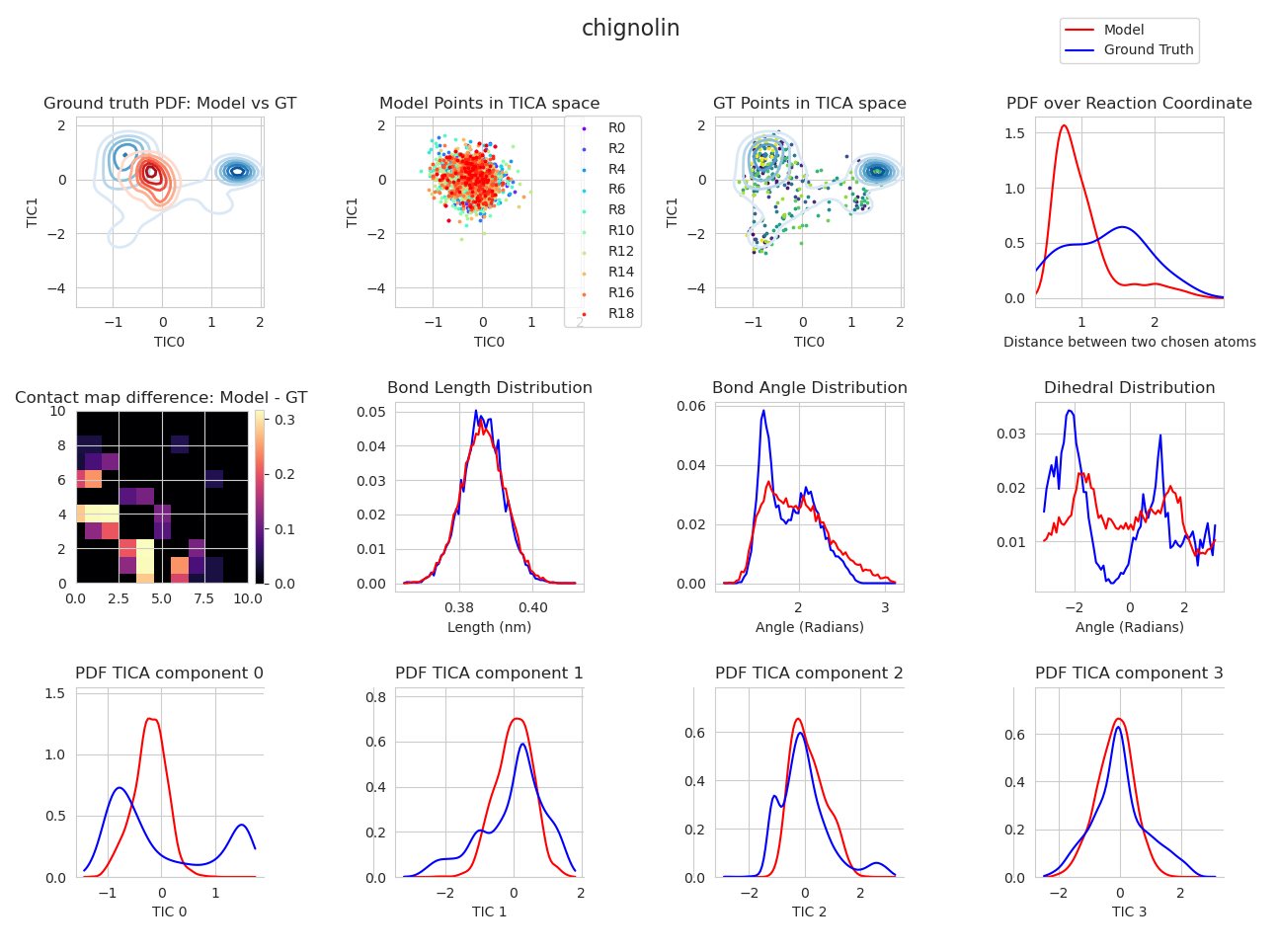}
    \caption{Benchmark suite evaluating the base model's performance after applying the active learning loop. Note: Without an unreasonable amount of computation time, the benchmark is unable to explore the entire space, so the first graph won't reach the entire ground truth distribution. However, the improvement is still visible, as the post-AL benchmark depicts a more broad distribution, implying more exploration and coverage.}
    \label{fig:post_compare}
\end{figure}

\end{document}